\documentclass[conference]{IEEEtran}
\IEEEoverridecommandlockouts
\usepackage{cite}
\usepackage{amsmath,amssymb,amsfonts}
\usepackage{algorithmic}
\usepackage{graphicx}
\usepackage{textcomp}
\usepackage{xcolor}
\def\BibTeX{{\rm B\kern-.05em{\sc i\kern-.025em b}\kern-.08em
    T\kern-.1667em\lower.7ex\hbox{E}\kern-.125emX}}
\usepackage{datetime}
\newdateformat{specialdate}{\twodigit{\THEDAY} \ \monthname[\THEMONTH] \THEYEAR}
\usepackage{subcaption}
\usepackage{grffile}
\usepackage{booktabs}
\usepackage{todonotes}
\usepackage{adjustbox}
\usepackage{amsmath,amssymb} % needed for math
\usepackage{array}
\usepackage{nicematrix}
\usepackage{multirow}
\usepackage{ntheorem}
\usepackage{csquotes}

\newcounter{hypocounter}
\renewcommand\thehypocounter{(H\arabic{hypocounter})}
\begin{document}

	\title{
		Evaluation of Orientation Ambiguity and Detection Rate in April Tag and WhyCode
	}

	\author{
		\IEEEauthorblockN{Joshua Springer}
		\IEEEauthorblockA{\textit{Department of Computer Science} \\
		\textit{Reykjavík University}\\
		Reykjavík, Iceland \\
		\texttt{joshua19@ru.is}}
		\and
		\IEEEauthorblockN{Marcel Kyas}
		\IEEEauthorblockA{\textit{Department of Computer Science} \\
		\textit{Reykjavík University}\\
		Reykjavík, Iceland \\
		\texttt{marcel@ru.is}}
	}

	\maketitle

	\begin{abstract}
	Fiducial systems provide a computationally cheap way for mobile robots
to estimate the pose of objects, or their own pose,
using just a monocular camera.
However, the orientation component of the pose of fiducial markers is unreliable,
which can have destructive effects in autonomous drone landing
on landing pads marked with fiducial markers.
This paper evaluates the April Tag and WhyCode fiducial systems
in terms of orientation ambiguity and detection rate on embedded hardware.
We test 2 April Tag variants -- 1 default and 1 custom --
and 3 Whycode variants -- 1 default and 2 custom.
%This informs subsequent drone landing tests that actuate a camera
%to track detected markers
%and use the marker's orientation in pose estimates.
We determine that they are suitable for autonomous drone landing applications
in terms of detection rate,
but may generate erroneous control signals
as a result of orientation ambiguity in the pose estimates.

%This paper evaluates April Tag and WhyCode fiducial systems in terms of 
%orientation ambiguity and detection rate,
%motivated by their application in autonomous drone landing.
%We propose 2 methods for mitigating the orientation ambiguity of WhyCode,
%and 1 method for increasing the runtime detection rate of April Tag.
%We evaluate our 3 systems against 2 default systems in terms of
%marker orientation ambiguity,
%and detection rate.
%We test rates of marker detection in a ROS framework on a Raspberry Pi 4,
%and we rank the systems in terms of their performance.
%Our first WhyCode variant reduces orientation ambiguity but does not increase detection rate.
%Our second WhyCode variant does not reduce orientation ambiguity,
%but provides the ability to track multi-marker WhyCode bundle arrangements.
%Our April Tag variant does not show performance improvements on a Raspberry Pi 4.

	\end{abstract}

	\begin{IEEEkeywords}
		fiducial, marker, orientation, ambiguity, pose
	\end{IEEEkeywords}

	\section{Introduction}
	Fiducial markers provide a computationally cheap way to estimate the pose
(position + orientation)
of objects,
or for a robot to estimate its own pose in an environment.
Given a monocular image and the distortion parameters of the camera that produced it,
a fiducial system can quickly determine if the image contains one of its markers,
and can determine the marker's pose.
While the position component of the pose is typically accurate,
the orientation is often subject to ambiguity
(see Section~\ref{section:related_work}),
such that it
(and the pose to which it belongs)
has discontinuities when viewed as a time series.
One way of evaluating the ability of a fiducial system
to accurately determine the orientation of the marker,
without having a sophisticated system to know the marker's ground-truth pose,
is to detect discontinuous poses perceived from fiducial markers
attached to physical objects.
Further, since mobile robots that can benefit from fiducial markers use
embedded hardware with limited computational capacity,
it is useful to know the rate at which the fiducial system can detect markers
when executing on such hardware.
Previous work tends to use a downward-facing camera that is
fixed on the drone or stabilized on a gimbal, but does not actuate in order to track the landing pad;
our evaluation informs subsequent autonomous drone landing tests,
where a drone actuates its camera
%-- in contrast to previous work that uses a fixed or stabilized, downward-facing camera --
in order to actively track a landing pad that is marked with a fiducial marker~\cite{spark_landing_paper_arxiv}.
The drone uses the marker's pose to generate position targets
in order to approach and land on the landing pad.
In this scenario, pose discontinuities caused by orientation ambiguity
propagate to the drone's control signals and cause erratic behavior.
Moreover, in such a time-sensitive scenario, the pose estimation must be fast,
motivating our evaluation of the systems' detection rates.

\begin{figure}[]
    \centering
    \begin{subfigure}[b]{0.45\linewidth}
        \includegraphics[width=\textwidth]{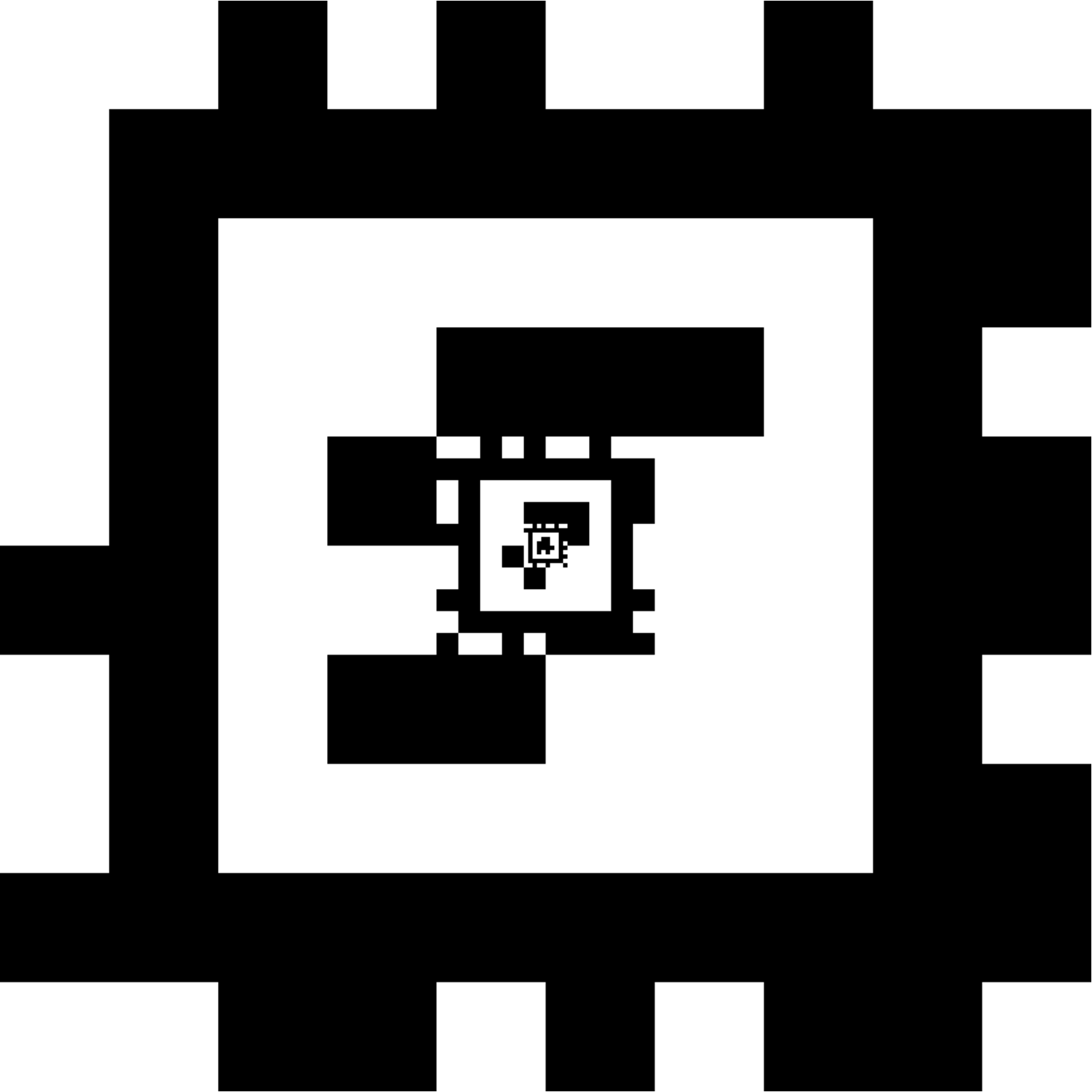}
        \caption{April Tag 48h12}
        \label{figure:apriltag48h12}
    \end{subfigure}
    \begin{subfigure}[b]{0.45\linewidth}
        \includegraphics[width=\textwidth]{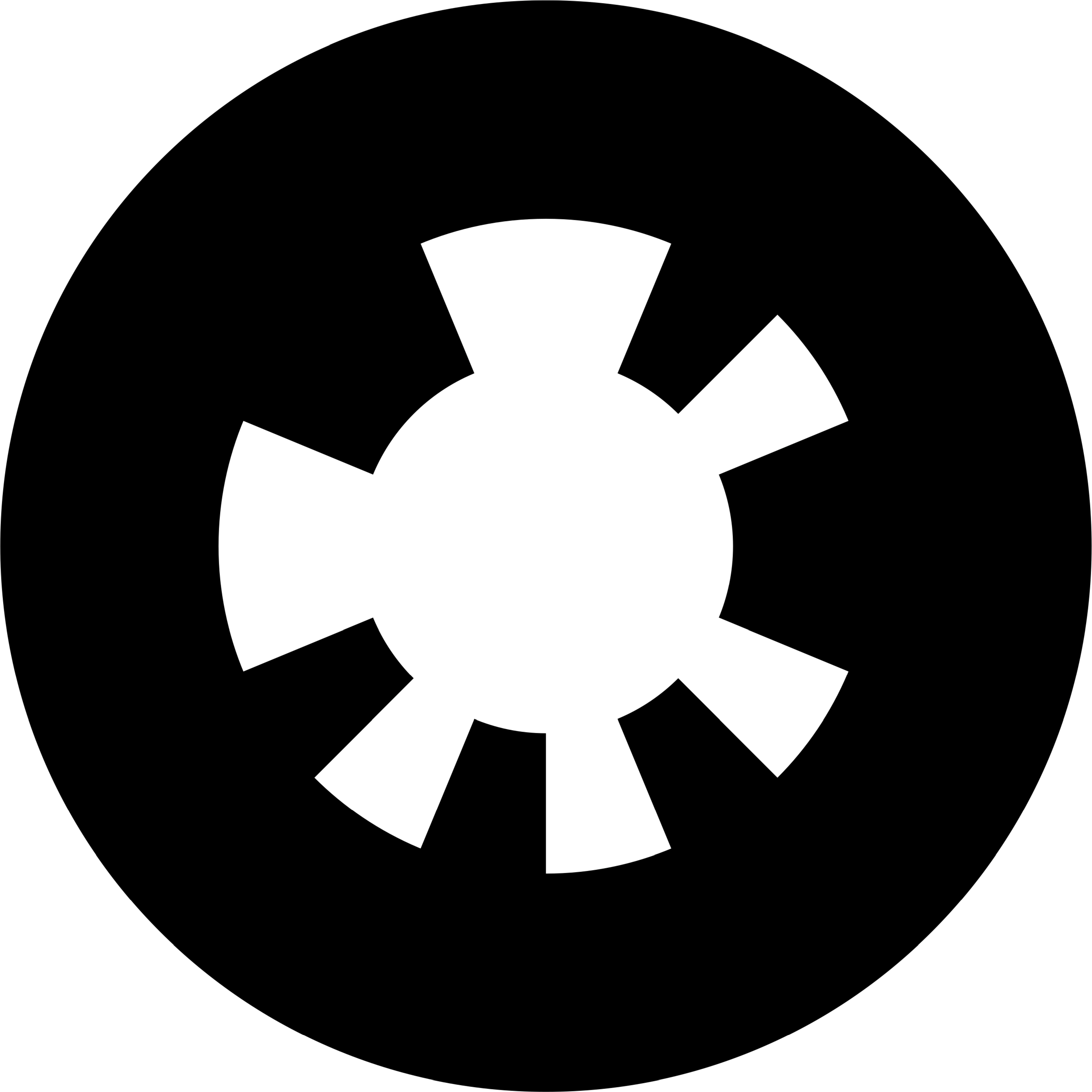}
        \caption{WhyCode}
        \label{figure:whycode_single}
    \end{subfigure}

    \begin{subfigure}[b]{0.45\linewidth}
        \includegraphics[width=\textwidth]{./images/tagCustom24h10_00002_00001_00000}
        \caption{April Tag 24h10}
        \label{figure:apriltag24h10}
    \end{subfigure}
    \begin{subfigure}[b]{0.45\linewidth}
        \includegraphics[width=\textwidth]{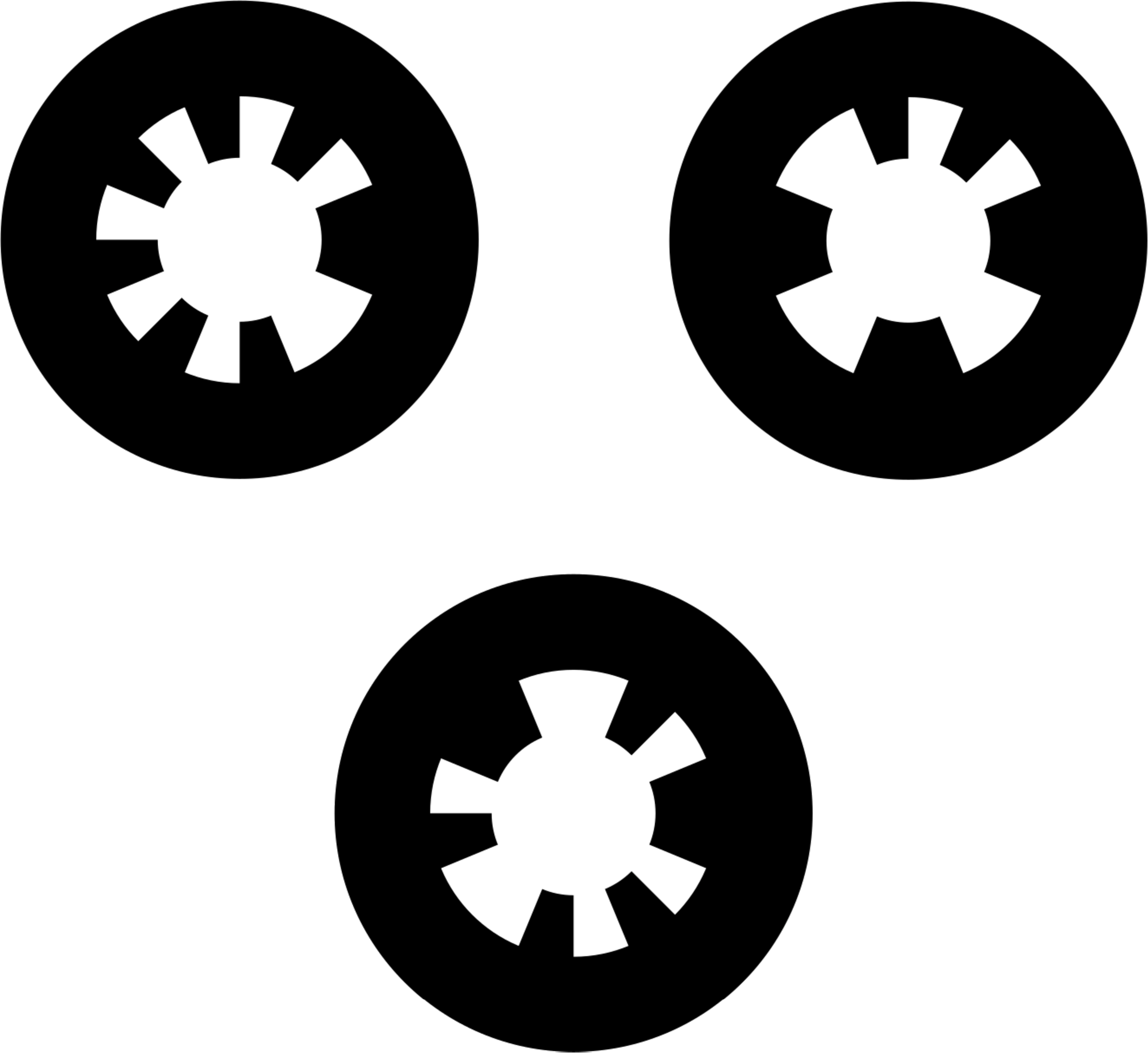}
        \caption{WhyCode ``Bundle''}
        \label{figure:whycode_bundle}
    \end{subfigure}
    \caption{The fiducial markers evaluated in this paper.}
    \label{figure:marker_setup}
\end{figure}

We evaluate a total of 5 fiducial systems,
2 of which are the default, unmodified systems of April Tag 48h12 and WhyCode
(Figures~\ref{figure:apriltag48h12} and~\ref{figure:whycode_single} respectively).
We also tested three modified versions of the default systems:
1) our ``WhyCode Ellipse'' variant which uses the same marker
as in Figure~\ref{figure:whycode_single} but uses additional image sampling points
to inform its decision of the marker's orientation,
2) our ``WhyCode Multi'' variant, which uses the marker arrangement in
Figure~\ref{figure:whycode_bundle} to determine the orientation of the plane connecting
the markers, and then assigns that orientation to all of the markers,
and 3) the April Tag 24h10 variant, a smaller version of April Tag 48h12.
We focus on monocular fiducial systems because
they are computationally cheap
and can therefore execute on embedded hardware onboard a drone,
and they are cheap to deploy, requiring only a printout of a particular marker,
and a monocular camera (arguably the most common drone peripheral sensor),
There are many possible fiducial systems
~\cite{ar_tag,aruco_orig,whycode_paper,whycon_paper,apriltag_paper},
but we choose those
that are highly configurable,
open source,
and whose positional accuracy has been formally evaluated.
Some systems that address the issue of orientation ambiguity
(such as Lentimark~\cite{lentimark} and the filtering method described in~\cite{chng2019resolving})
involve fundamental changes to the marker systems
or are proprietary and are, therefore, less widely used.
We aim to produce simpler solutions requiring minimal changes to existing marker systems to allow easy integration.

	\section{Background}
	\label{section:background}

Fiducial markers are 2-dimensional patterns whose positions (and IDs)
can accurately be determined in space using only images, without a notion of time.
WhyCode~\cite{whycode_paper} (shown in Figure~\ref{figure:whycode_single}
and~\ref{figure:whycode_bundle})
is a lightweight, circular fiducial marker system, formed by an inner white circle, outer black circle, and a Manchester encoded ID between the two circles.
April Tag~\cite{apriltag_paper,apriltag2_paper,apriltag3_paper}
(shown in Figures~\ref{figure:apriltag48h12} and~\ref{figure:apriltag24h10})
is a square marker system with a configurable layout white and black squares,
each representing a data bit or a section of the black and white border.

It is difficult to determine the orientation of planar fiducial markers using single-image, monocular systems,
due to the fact that a marker may have the same appearance when viewed from specific, different angles.
This orientation ambiguity typically results in apparent discontinuities in a marker's orientation when it is viewed as a time series.
If the full marker pose (position + orientation) is used as a control input to the drone, these discontinuities can cause potentially destructive, erratic behavior.
Therefore, it is easier to achieve stable precision drone landing using a fixed, downward-facing camera,
so that the control algorithm needs only to consider the position of the fiducial marker and can ignore its orientation.
Conversely, a marker system that avoids these discontinuities would enable gimbal-based marker tracking and therefore more reliable precision landing.

	\section{Related Work}
	\label{section:related_work}

%Several projects have accomplished autonomous precision drone landing with fiducial markers and a \textit{fixed} camera.
%Wynn~\cite{wynn} uses ArUco markers to land on a ship after an initial GPS-based approach,
%with a smaller marker nested inside a larger marker for more detection range.
%Borowcyk et~al.~\cite{high_velocity_landing} use a single April Tag marker, with multiple GPS receivers and data links
%to land a DJI Matrice on a moving ground vehicle.
%Falanga et~al.~\cite{vision_based_x_platform} use 2 fixed cameras and a specially designed marker to land a drone on a moving platform indoors.
%Wubben et~al.~\cite{accurate_landing_UAV_ground_pattern} use a single downward-facing camera
%and 2 non-nested ArUco~\cite{aruco_orig} markers to identify and land on a landing pad,
%reporting some losses in marker detection as a result of wind pushing the drone during landing.
%In the above projects, the fixed camera allows the landing system to consider only the position of the marker's pose (since the orientation of the camera is known), and not its orientation, thereby eliminating the problem of orientation ambiguity.
%However, the fixed camera also makes it more difficult to track the markers,
%since multi-rotor drones change orientation in order to vector their thrust.
%We evaluate the orientation ambiguity of our systems to determine if they are applicable in a gimbaled setting to allow for marker tracking,
%thereby increasing reliability of the landing method.

Irmisch~\cite{apriltag_whycon_comparison} has analyzed the performance of April Tag and WhyCon markers in terms of their abilities to determine
correct position in real world experiments.
The experiments show that both April Tag and WhyCon have relatively low position estimate errors.
The issue of ambiguity in planar marker orientation is a known problem, and various methods exist which attempt to mitigate it.
One example is an edition of ARTag called LentiMark~\cite{lentimark},
which uses special Moir\'e patterns on the outside of the tags to determine a correct orientation.
Another study~\cite{chng2019resolving} proposes to determine the correct solution using an averaging algorithm with multiple views of the markers,
instead of using only a single image.
The method in~\cite{planar_pose_ambiguity_motion_model} proposes to consider the marker as a moving object,
and therefore reduces orientation ambiguity by intelligently choosing the orientation using a motion model.

	\section{Methods}
	\label{section:methods}

We captured the video stream at 480p with a \enquote{Creative Technology Live! Cam Sync 1080p.}
The camera has a wide angle of 77 degrees.
We mounted the markers to a divider wall and moved the camera by hand to simulate the drone's approach,
and tested at distances of 1 to 3 meters,
as relevant in the subsequent autonomous drone landing scenario~\cite{spark_landing_paper_arvix}.
We performed the experiments in a lab room at Reykjavik University,
allowing us to control the light,
whereas outdoor environments have variable light conditions and spectral signatures~\cite{hedgecoe_2003}.
We use the position targets to evaluate the systems' performance
in terms of the reliability of the control signals they
would send to a drone (orientation ambiguity),
and at what frequency the system recognizes the landing pad at all (detection rate).
%The camera fails to recognize the markers at XX m. todo!!! (if the paper gets accepted.
% 							     ROS is not cooperating right now
%							     so I will try later.)

\subsection{ROS Message Attributes and Calculations}

April Tag and WhyCode both have ROS (Robot Operating System)~\cite{ros} modules
that allow them to interact with other programs, such as flight control software.
We have extended them to perform calculations and include message attributes for drone landing:
\begin{enumerate}
    \item A \textbf{position target} in the a relative ``east, north, up'' (ENU)
	    coordinate frame \emph{whose origin is the camera},
	    where ``east'' refers to the camera's right, ``north'' to the front,
		and ``up'' to the top.
	  This parameter gives the relative position from the camera to the marker,
	  under the assumption that the marker is flat on the ground, facing up.
	  In an autonomous drone landing scenario, this parameter tells the drone
		where to go in order to approach the marker.
    \item The \textbf{normalized pixel position} $u_n, v_n \in [-1,1]$ of the center of each marker, which serves as inputs to the systems that track the marker.
    \item The marker's \textbf{orientation components}: yaw, pitch, and roll.
          Our edits exposed these to the ROS interfaces in the cases where they were not explicitly exposed.
          An autonomous drone can use the marker's yaw
		to align itself to the landing pad before descent.
\end{enumerate}

\subsection{Tested Fiducial Systems}
\label{section:tested_fiducial_systems}

\subsubsection{WhyCode Orig}
\label{section:whycode_orig}
We use a version of WhyCode created by Ulrich~\cite{ulrich} as a baseline for testing.
The method samples the ellipse that goes through the centers
of the ``teeth'' forming the marker's ID
(i.e.~the yellow and green ellipses in
Figure~\ref{figure:whycode_ellipse_both_solutions_cropped}),
to determine the marker's orientation.
Of the two possible candidate solutions that are implied by the detected semi-axes of the outer black ellipse, the detector chooses the one with a lower variance in the number of sample points per tooth.
This works because the candidate solutions predict this ellipse to be in slightly different places,
and the correct solution should predict the ellipse to be in its correct place,
minimizing the variance in the number of sample points that coincide with each tooth.
Conversely, the incorrect solution should predict the ellipse to be in the incorrect place,
such that the sampling ellipse does not line up well with the marker, and the variance is higher.
We use WhyCode markers with 8-bit IDs so that there are several sample points and a meaningful variance.

\begin{figure}
    \centering
    \includegraphics[width=0.7\linewidth]{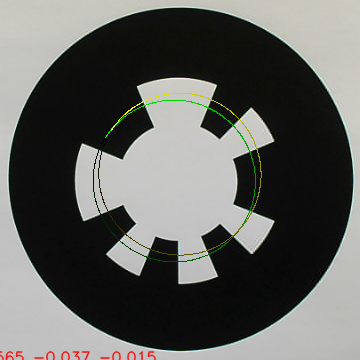}
    \caption{An illustration of the method in~\cite{ulrich} that determines an orientation of the marker based on
    which ellipse is better centered on the marker. It chooses the ellipse that minimizes the variances in the ellipse's intersection with the teeth.
    Both possible solutions are shown, with the green being correct.}
    \label{figure:whycode_orig_both_detections}
\end{figure}

\subsubsection{WhyCode Ellipse}
The first method that we have implemented for reducing orientation ambiguity is the \texttt{ellipse\_sampling} branch of~\cite{uzgit_whycon}.
The method determines the marker ID and the two candidate solutions for the orientation as in WhyCode Orig,
after which it identifies the lines that go from the center of the white region through the center of each tooth.
It then samples the input image on these lines,
as illustrated by Figure~\ref{figure:whycode_ellipse_both_solutions_cropped}.
It expects a white-to-black transition at the predicted edge of each tooth, and the sampling line is centered on this edge.
The true edge is determined during sampling, and its value is recorded as a percentage of the length of the line segment,
oriented such that 0 corresponds to the centermost end of the line segment, and 1 corresponds to its outermost end.
The detector chooses the solution that minimizes the variance of the location of the true edge over the sample lines.

\begin{figure}
    \centering
    \includegraphics[width=0.7\linewidth]{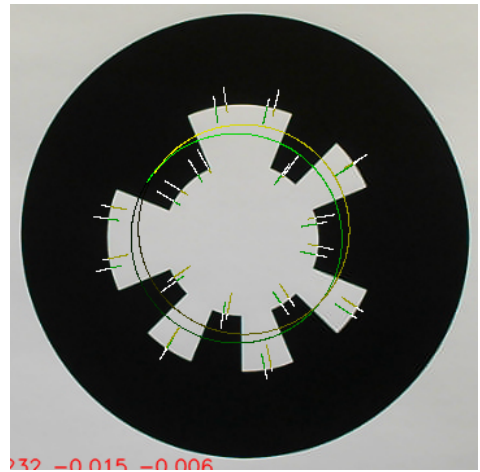}
    \caption{An illustration of ``WhyCode Ellipse''.
	Each ellipse and corresponding radial lines represent the sampling locations,
	 marked in yellow and green to distinguish the two candidate solutions.
	This method uses the radial lines to determine how well each 
	candidate solution is aligned with the real image.
	In this case, the green solution is correct and is visually better centered on the marker.}
    \label{figure:whycode_ellipse_both_solutions_cropped}
\end{figure}

\subsubsection{WhyCode Multi}
The second method that we have implemented for reducing orientation ambiguity
(the \texttt{multi} branch of~\cite{uzgit_whycon})
works under the assumption that all recognized markers are coplanar.
For each input image, the WhyCode algorithm identifies all markers and
then finds the normal vector to the plane implied by the markers' positions,
after which it can calculate the pitch and roll components of the bundle's orientation.
The position of the bundle is the mean of the positions
of its constituent markers,
and its yaw is that of any constituent markers, with the assumption they are all the same.
The detector then calculates all additional attributes for the bundle
as if it were a single marker.
This system uses the arrangement of WhyCode markers in Figure~\ref{figure:whycode_bundle}.

\begin{figure}
	\centering
	\includegraphics[width=\linewidth]{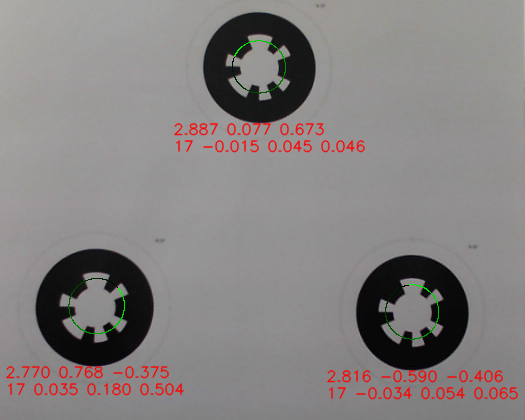}
	\caption{An illustration of ``WhyCode Multi''. The system regresses a plane connecting the markers, then assigns that orientation to each marker, as they are assumed to be coplanar.}
	\label{figure:whycode_ellipse_both_solutions_cropped}
\end{figure}

\subsubsection{April Tag 48h12}
April Tag provides a default family 48h12, shown in Figure~\ref{figure:apriltag48h12},
for which the 4 center squares are undefined,
20 squares provide a white border,
28 squares provide a black border,
and 48 squares provide data bits,
giving a total of 96 defined squares.
The undefined center provides a space to embed a smaller marker,
which is useful in the last stages of a drone landing scenario,
where the camera is too close to the landing pad to see the larger markers.
We test this family as a baseline for the performance of April Tag.

\subsubsection{April Tag 24h10}
April Tag provides libraries to specify and generate non-default marker variants,
from which we generated the April Tag 24h10 family
(shown in Figure~\ref{figure:apriltag24h10}).
%in response to our initial experiments, which showed that April Tag 48h12 has a slower detection rate than WhyCode on embedded hardware.
This variant maintains marker embeddability
while decreasing the size of the marker definition,
by reducing the size of the undefined center to one square,
and adjusting the surrounding regions accordingly:
Eight squares for a white border, 16 squares for a black border, and 24 outer data squares.
This gives a total of 48 defined squares(compared to the 96 squares of April Tag 48h12).
This reduction is important because all possible valid marker IDs for each family
are loaded into a single hash table at runtime, and April Tag 48h12
has a large hash table of 42,211 markers which can require more than 1 GB of RAM
-- a precious resource on embedded hardware.
By contrast, April Tag 24h10 has only 18 markers and therefore requires significantly less memory.
We test April Tag 24h10 to see if it can offer an increase in detection rate
with respect to April Tag 48h12.

\subsection{Bases of Comparison}

\subsubsection{Discontinuities}
Orientation ambiguity manifests as discontinuities (e.g.~sign flips and other spikes)
in the pitch and roll components of a marker's orientation,
which propagate to subsequent calculations that depend on the orientation
-- \textit{the position targets}.
The fiducial systems are compared on the basis of
the number of discontinuities in the position targets they generate.
A ``good'' system minimizes this number.

We capture 33 videos of the marker arrangement in Figure~\ref{figure:marker_setup},
printed, so each marker has a side length of 30 cm,
and mounted to a planar surface with clear lighting.
We move the camera in each of the videos (e.g.~panning, tilting, moving in and out, etc.)
while keeping all markers completely in the frame at all times.
We save the videos as a series of pairs of image and camera info messages
in the standard ROS way, using \texttt{rosbag}.
Each fiducial system in Section~\ref{section:tested_fiducial_systems}
processes the same set of videos.
Since the experiments are conducted slowly in a controlled environment,
angular and linear speeds above pre-determined thresholds
can be classified as discontinuities.
Linear discontinuities occur when the quotient of any position target
$\vec{P} = \langle p_e, p_n, p_u \rangle$ (east, north, or up)
and its predecessor is sufficiently negative
(such discontinuities always occur over the origin):
\begin{equation}
    \frac{p_{x,i+1}}{p_{x,i}} < \theta_l < 0
    \label{equation:linear_discontinuity}
\end{equation}
where $p_{x,i}$ and $p_{x,i+1}$ are position targets in a single dimension
$x \in \{e, n, u\}$ dimension at frames $i$ and $i+1$ respectively,
and $\theta_l < 0$ is an experimentally determined threshold
(see Table~\ref{table:discontinuity_thresholds}).
If the inequality in Equation~\ref{equation:linear_discontinuity} is true,
this implies that the marker appears to change locations
faster than allowed during testing.

Similarly, a discontinuity determined from angular speed $s_a$ occurs when
\begin{equation}
    s_a = \frac{\mathrm{dist} \left( q_i, q_{i+1} \right) }{\Delta t} > \theta_a > 0
    \label{equation:angular_discontinuity}
\end{equation}
    where $q_i$ and $q_{i+1}$ are the quaternions representing the orientation of a marker at frames $i$ and $i+1$, respectively,
$\mathrm{dist} \geq 0$ is the intrinsic geodesic distance between the angles represented by the quaternions,
$\Delta t$ is the change in time between frames $i$ and $i+1$,
and $\theta_a$ is a pre-determined threshold
(see Table~\ref{table:discontinuity_thresholds}).
If the inequality in Equation~\ref{equation:angular_discontinuity} is true,
this implies that the marker appears to rotate faster than allowed during testing.

We consider cases where
Equations~\ref{equation:linear_discontinuity}
and~\ref{equation:angular_discontinuity}
are true simultaneously,
to reduce false identification of discontinuities.
For example, a linear discontinuity can appear erroneously via noise
when the $e$ or $n$ component of a marker's position target is close to 0.
%and
%a discontinuity in angular speed can be caused by quirks in the notation
%of the orientation of a marker,
%when no linear discontinuity occurs (e.g.~jumping from $2\pi$ to 0 as a result of normalization).
If both discontinuities occur at the same time, this means that the marker truly appears
to drastically change positions in space in a small amount of time,
which does not happen in our test cases.
Finally, since the test cases vary in length,
we can define a ``discontinuity rate'' $r_d = \frac{d}{n}$
which describes the number of discontinuities $d$
as a portion of the total number of detections $n$,
and which serves as a basis
for comparing the performance of each system in each test case.

\subsubsection{Detection Rate}

We capture 14 videos of each marker in Figure~\ref{figure:marker_setup}
separately with both the camera and marker remaining still,
at several distances and deflections from one another.
We record videos of each marker in isolation to avoid any potential interference,
since the marker detectors all analyze black and white regions.
Both the camera and marker remain still during each test case,
and each test case lasts 60 s.
The systems are compared on the basis of their detection rate~(Hz)
when executing on a Raspberry Pi 4 with 2 GB of RAM,
with the following ROS pipeline:
replaying a test case video,
image rectification using \texttt{image\_proc}
(April Tag systems only, required by April Tag ROS module),
marker detection by each of the fiducial systems
in Section~\ref{section:tested_fiducial_systems},
and rosbag recording for later analysis.
This avoids capturing the image during testing,
but maintains fairness of comparison among the systems,
since all systems are put into the same computational environment one at a time,
and all process similar test cases.
We then determine a detection rate $F=\frac{n}{t}$, where $n$ is the number of detections, and $t$ is the length in seconds of the test case.
The goal of this metric is to determine at what frequency the embedded hardware can
run the fiducial system.
All systems process video from the same camera at the same framerate,
so the setup is not biased in favor of any particular system.

We use the default system parameters for both April Tag and WhyCode.
We also test only with the markers in Figure~\ref{figure:marker_setup},
as these are the markers that we use to mark landing pads.

	\section{Results}
	\label{section:results}

\begin{figure}[]
	\centering
	\includegraphics[width=0.95\linewidth]{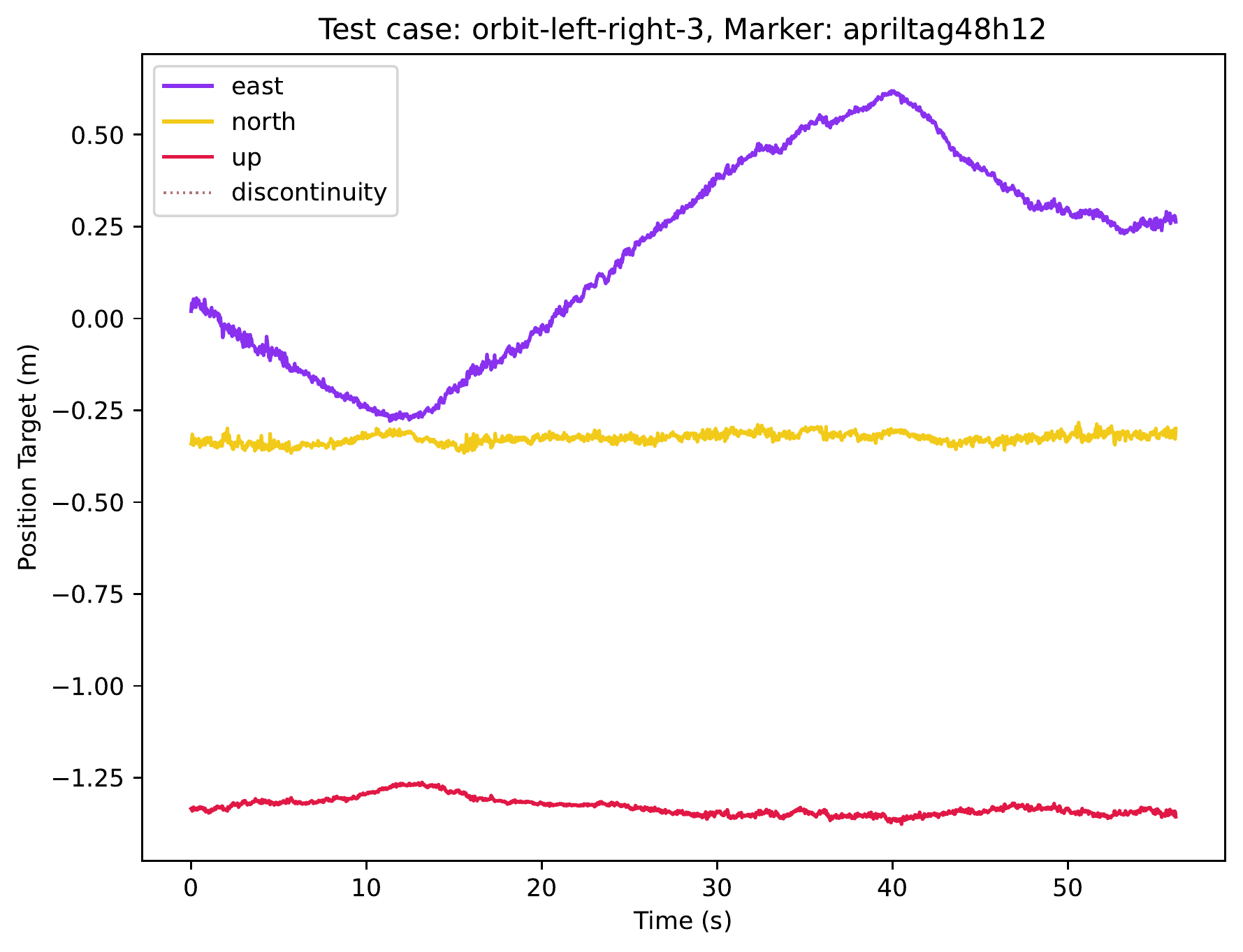}
	\caption{Example of position targets without discontinuities.}
	\label{figure:non_discontinuity_example}
\end{figure}

\begin{figure}[]
	\includegraphics[width=0.95\linewidth]{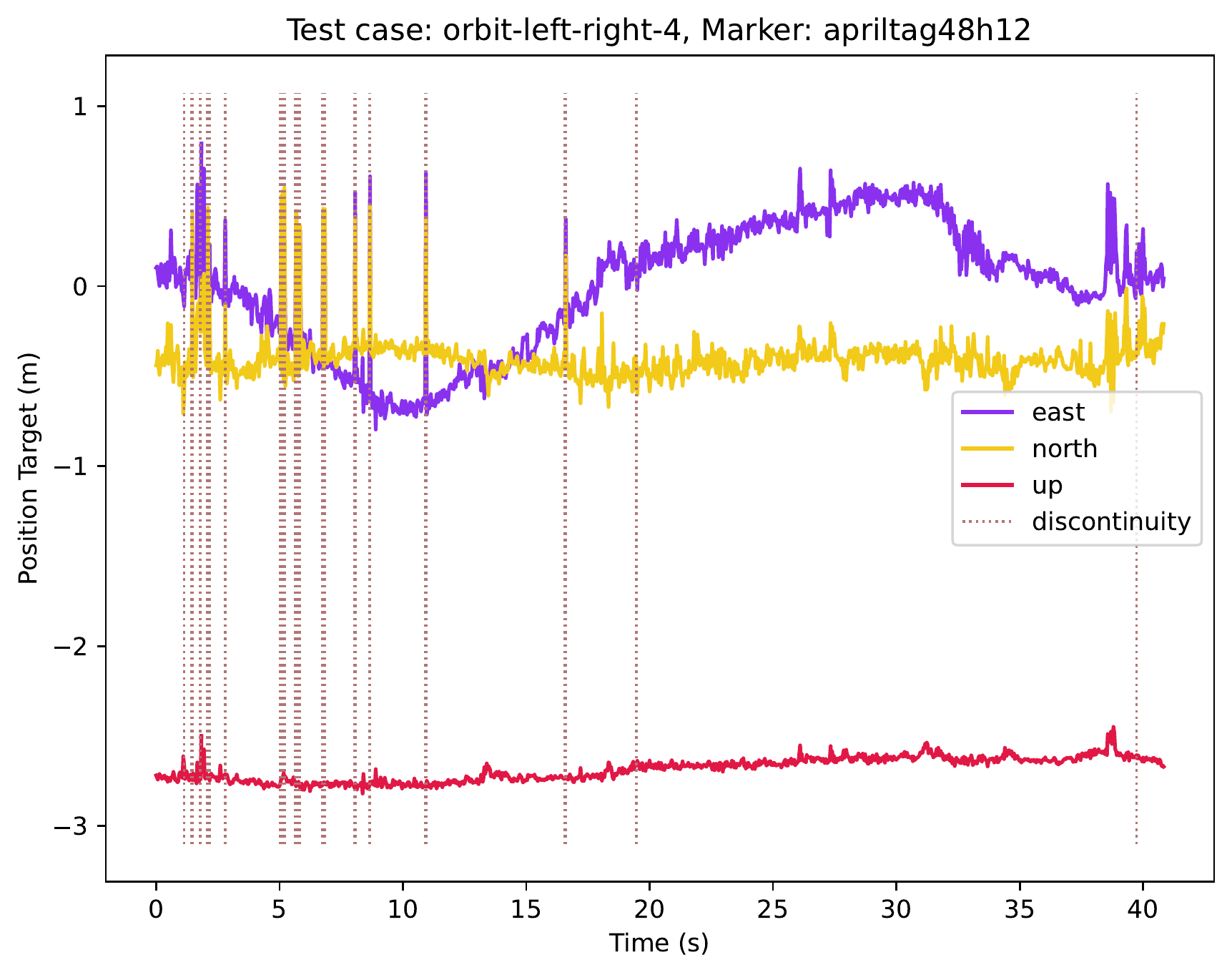}
	\caption{Example of discontinuities in position targets (marked by the vertical lines), which are spikes in the east and north position targets that roughly correspond to sign flips.}
	\label{figure:discontinuity_example}
\end{figure}

\begin{figure}[]
	\includegraphics[width=0.95\linewidth]{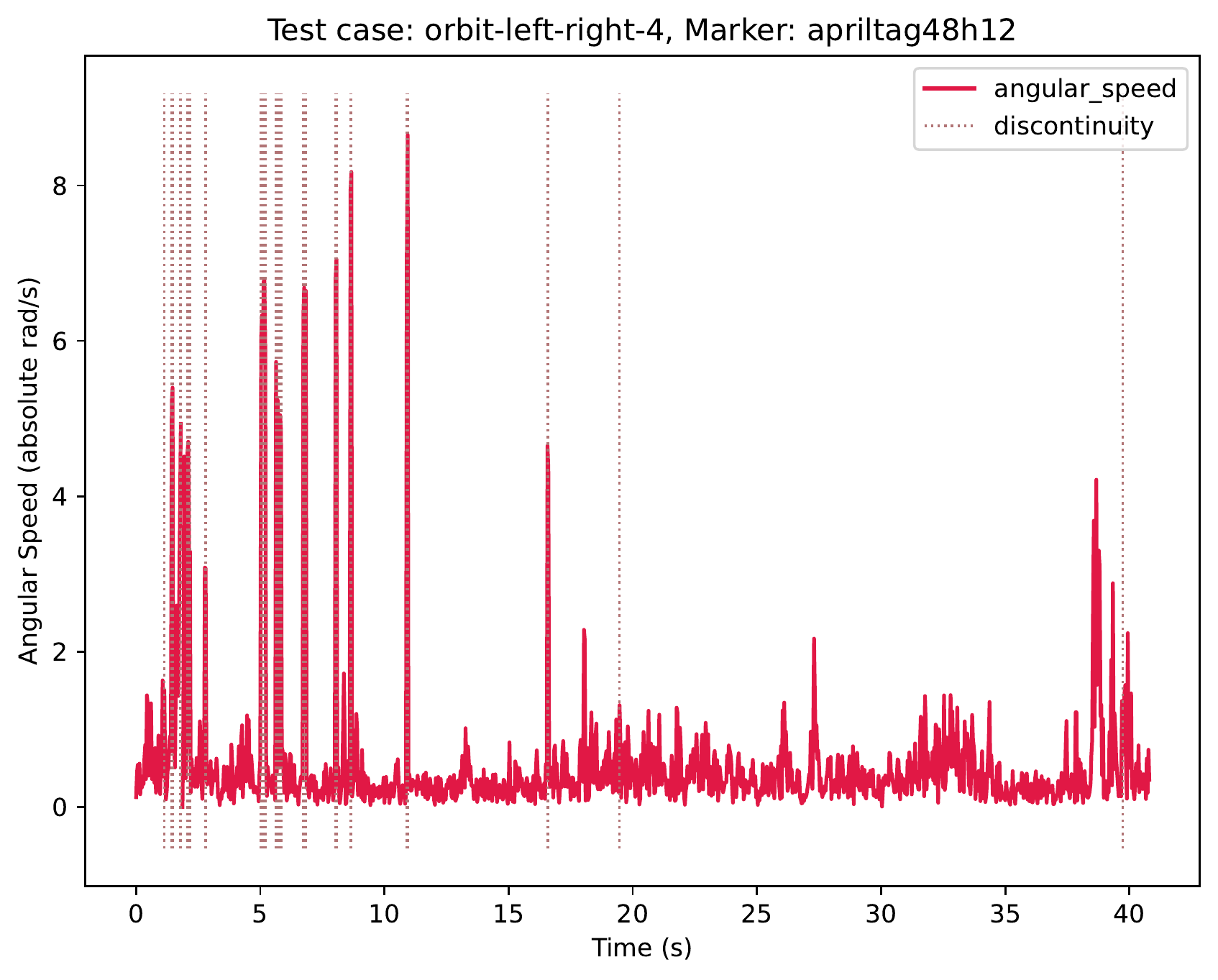}
	\caption{The angular speeds of the marker in Figure~\ref{figure:discontinuity_example}, similarly labeled with discontinuities.}
	\label{figure:discontinuity_example_angular_velocity}
\end{figure}

\begin{figure}[]
    \centering
    \includegraphics[width=\linewidth]{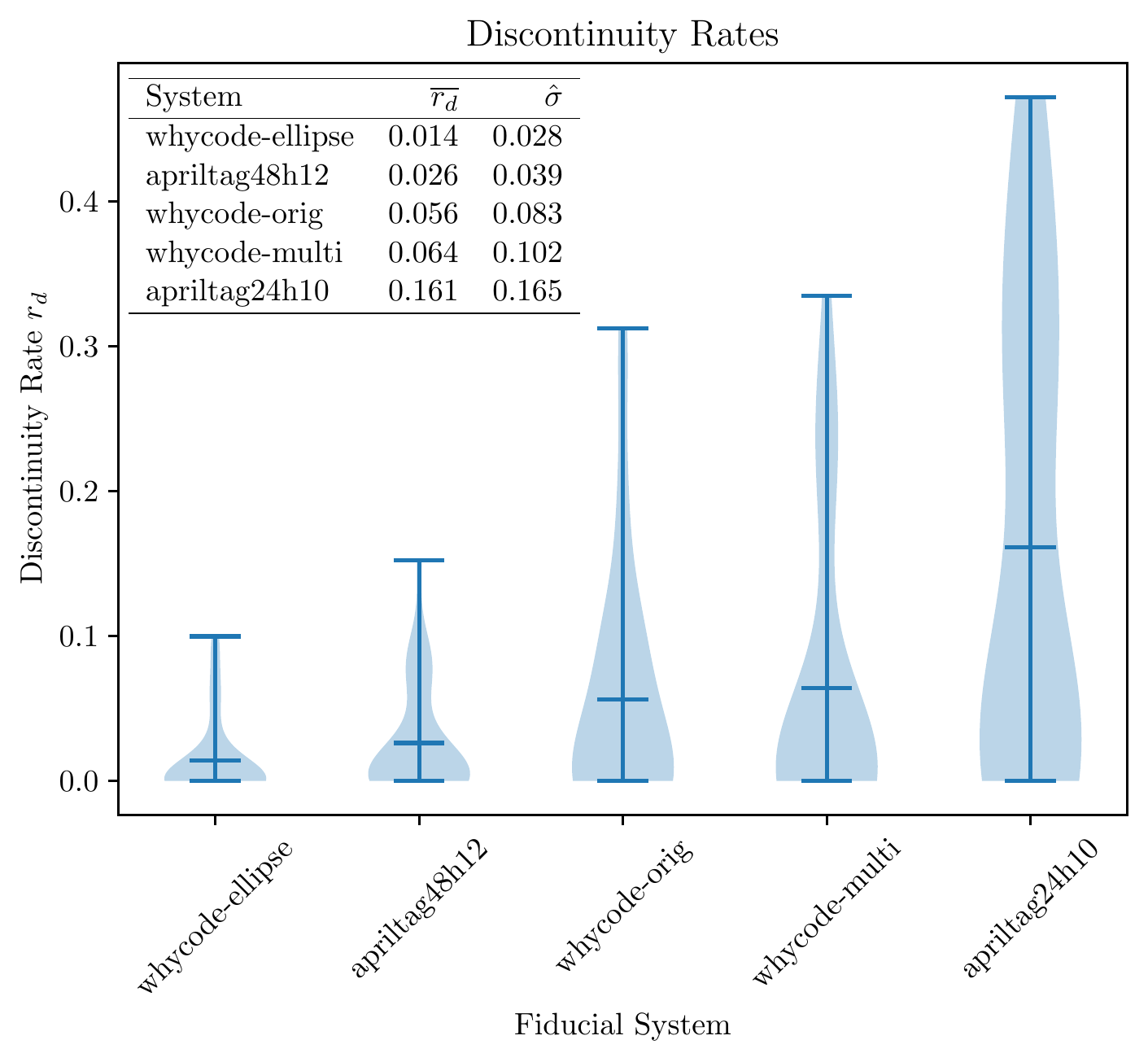}
    \caption{A visualization of the discontinuity rates $r_d$, which represent the proportion of discontinuous position target readings relative to the total number of readings. $\overline{r_d}$ is the sample mean of the discontinuity rate, and $\hat{\sigma}$ is the sample standard deviation, and $n=33$ per group.}
    \label{figure:violin_plot_five_member}
\end{figure}

\begin{figure}[]
    \centering
    \includegraphics[width=\linewidth]{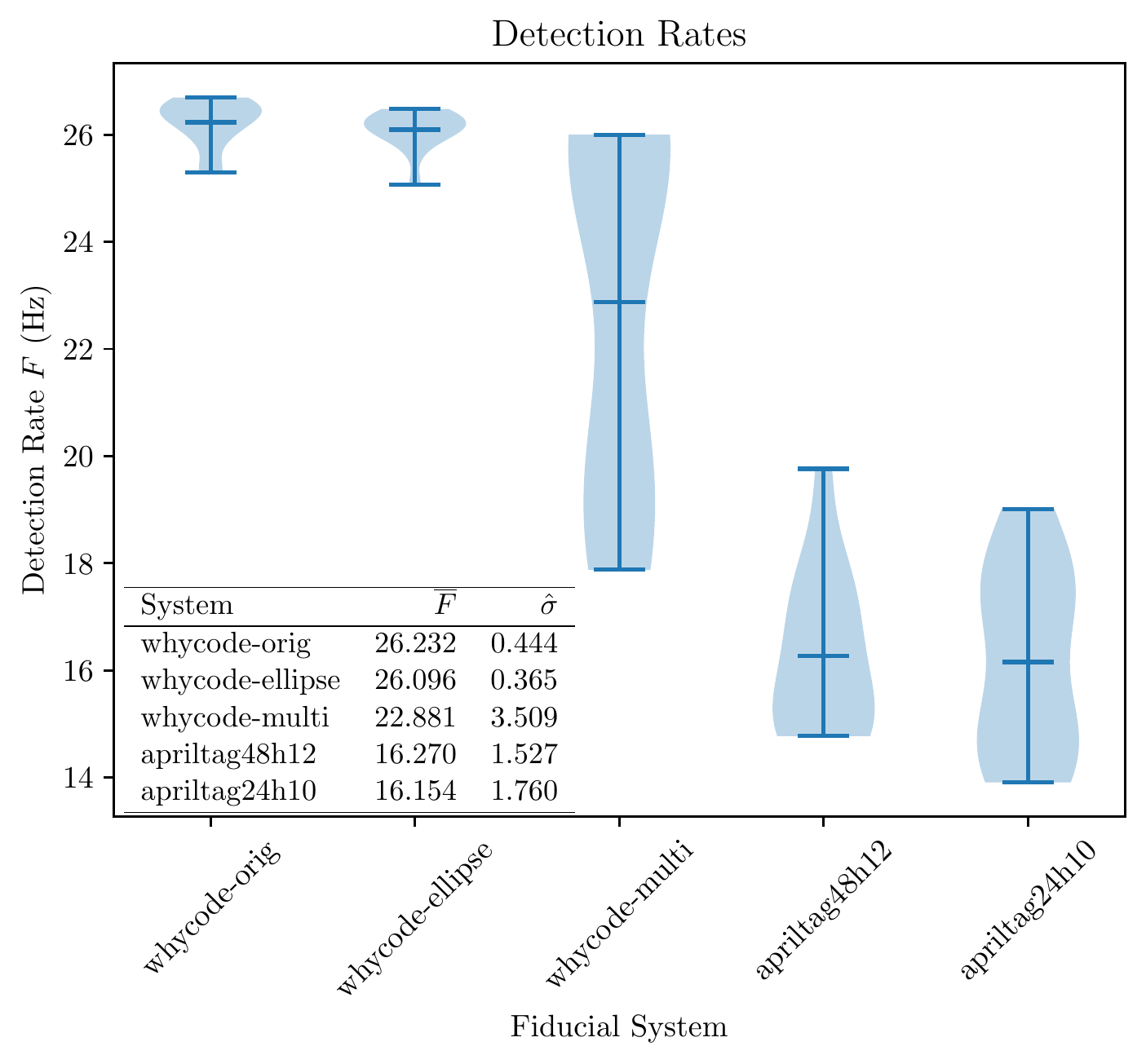}
    \caption{A visualization of the detection rates $F$ (Hz). $\overline{F}$ is the sample mean of the detection rate, and $\hat{\sigma}$ is the sample standard deviation, and $n=14$ per group.}
    \label{figure:violin_plot_speed_five_member}
\end{figure}

Figures~\ref{figure:non_discontinuity_example} and~\ref{figure:discontinuity_example}
give an intuition for the discontinuities studied in this paper.
Figure~\ref{figure:non_discontinuity_example} shows the east, north, and up position targets
for a test case where the camera moves first to the left, then to the right,
keeping the marker near the center of the frame the entire time -- ``orbiting'' the marker.
The east position target correctly indicates the camera's movement left and right,
while the north and up position targets correctly remain near-constant
(the camera was being moved by hand, so there is some erroneous movement).
Figure~\ref{figure:discontinuity_example} shows the next test case with the same marker
and same camera movement, but from a longer distance (indicated by the lower up position target).
Since the marker is farther away and therefore smaller in the camera frame,
it is more difficult for the system to estimate its orientation,
leading to the discontinuities.
The discontinuities always occur over the origin --
the camera appears to jump from one side of the marker to the other.
Notice the spikes in the east position target $t\in[25,30]$s in Figure~\ref{figure:discontinuity_example}
are not discontinuities, but rather noise.
Further, noise in the east position target in both
Figures~\ref{figure:non_discontinuity_example} and~\ref{figure:discontinuity_example}
would be considered discontinuities when the position target becomes near-zero,
such that noisy movement actually satisfies Equation~\ref{equation:linear_discontinuity}.
Figure~\ref{figure:discontinuity_example_angular_velocity} shows the detected angular speed
of the marker in Figure~\ref{figure:discontinuity_example_angular_velocity},
showing that certain spikes in Figure~\ref{figure:discontinuity_example} are not considered
discontinuities because they do not correspond in time to spikes in angular speed.
Figure~\ref{figure:violin_plot_five_member} visualizes the discontinuity rates for the systems,
and Figure~\ref{figure:violin_plot_speed_five_member} visualizes their detection rates.
Table~\ref{table:discontinuity_thresholds} shows the speed thresholds for Equations~\ref{equation:linear_discontinuity} and~\ref{equation:angular_discontinuity}.

\begin{table}[]
    \centering
    \begin{tabular}{lll}
\toprule
Discontinuity Type &     Symbol &           Value \\ \hline
% \midrule
        rotational & $\theta_a$ &       1.0 rad/s \\
            linear & $\theta_l$ & -0.8 (unitless) \\
\bottomrule
\end{tabular}

    \caption{Thresholds for targeting pose discontinuities in all systems. We say that a discontinuity occurs when the inequalities in Equations~\ref{equation:linear_discontinuity}
    and~\ref{equation:angular_discontinuity} are simultaneously true using the values in this table.
    These are chosen to be well above the maximum allowed values in testing, so that they are not sensitive to noise.
    }
    \label{table:discontinuity_thresholds}
\end{table}

	\section{Discussion}
	WhyCode Ellipse offers a decrease in discontinuity rate compared to WhyCode Orig,
with only a small decrease in detection rate.
WhyCode Multi does not offer a decrease in discontinuity rate compared to WhyCode Orig,
and also has a lower detection rate.
It is possible that variations in the orientation of the plane connecting
WhyCode Multi markers could be reduced by adding more markers to the arrangement,
or spreading them out.
%However, it shows that the notion of a marker ``bundle'' can be extended to WhyCode,
%and this opens up possibilities for pseudo-embedded marker arrangements.
April Tag 24h10 has a higher discontinuity rate than April Tag 48h12,
with no increase in detection rate.
The lower detection rates of the April Tag systems
are likely a result of their longer ROS pipeline,
which depends on transmitting each input image to an intermediate \texttt{image\_proc} node and back for rectification.
WhyCode builds this into its algorithm, thereby avoiding latency and performing faster than April Tag even when detecting several markers at a time.
However, WhyCode systems cannot handle marker embedding as April Tag systems can.

We use only the default system parameters, and a single hardware setup.
Different setups will have different results,
e.g.~a camera with a telephoto lens might provide more accuracy at long distances,
but would require calibration at many different focal lengths.
Different computational hardware will also give different detection rates.

	\section{Conclusion \& Future Work}
	We have evaluated 5 fiducial systems - 2 existing variants: April Tag 48h12 and WhyCode Orig,
and 3 custom variants which we have implemented - in terms of their rates of discontinuity in position target generation
and detection rate on a Raspberry Pi 4 (2 GB RAM).
We have determined that WhyCode Ellipse, WhyCode Multi, April Tag 48h12, and WhyCode Orig provide
a good starting point for testing gimbal-based fiducial landing with a drone,
while April Tag 24h10 is likely to exhibit problematic behavior.
These markers are further evaluated in our subsequent study~\cite{spark_landing_paper_arxiv}.

Future tests of the WhyCode Multi system could use different marker arrangements,
such as with more markers, or with the markers spaced farther apart.
All systems could be further tested
with different runtime parameters, cameras,
computational hardware, and lighting conditions.

%	\section{Future Work}
%	\input{sections/future_work.tex}

	\bibliography{references}
	\bibliographystyle{IEEEtran}

\end{document}